\documentclass{article}
\usepackage{spconf,amsmath,graphicx}
\usepackage{color}
\usepackage{graphicx}
\usepackage{float}
\usepackage{subfigure}
\usepackage{booktabs}
\usepackage{multirow}
\usepackage{setspace}

\title{Improving Retrieval-based Dialogue System via Syntax-Informed Attention}
\name{Tengtao Song, Nuo Chen, Ji Jiang, Zhihong Zhu, Yuexian Zou$^*$}
\address{ADSPLAB, School of ECE, Peking University, Shenzhen, China}
\begin{document}
\setlength{\textfloatsep}{5pt}

\maketitle
\begin{abstract}
Multi-turn response selection is a challenging task due to its high demands on efficient extraction of the matching features from abundant information provided by context utterances. Since incorporating syntactic information like dependency structures into neural models can promote a better understanding of the sentences, such a method has been widely used in NLP tasks. Though syntactic information helps models achieved pleasing results, its application in retrieval-based dialogue systems has not been fully explored. Meanwhile,  previous works focus on intra-sentence syntax alone, which is far from satisfactory for the task of multi-turn response where dialogues usually contain multiple sentences. To this end, we propose SIA, Syntax-Informed Attention, considering both intra- and inter-sentence syntax information. While the former restricts attention scope to only between tokens and corresponding dependents in the syntax tree, the latter allows attention in cross-utterance pairs for those syntactically important tokens. We evaluate our method on three widely used benchmarks and experimental results demonstrate the general superiority of our method on dialogue response selection.
\end{abstract}
\section{Introduction}
\label{sec:intro}

The past years have seen intensive research interests from academia and industry in the field of neural dialogue systems. Among mainstream types of dialogue systems, retrieval-based dialogue systems, such as Microsoft XiaoIce \cite{xiaoice} and Alibaba AliMe Assist \cite{ali}, that involve choosing optimal response from a set of candidates pool are widely used due to its rich response informativeness.

Most recently, large-scale pre-trained language models (e.g., BERT \cite{bert}, RoBERTa\cite{roberta}) grow in popularity and have shown superiority across a wide range of NLP tasks\cite{DBLP:conf/naacl/YouCLGWZ22, chen2022would}. In particular, BERT-based approaches have been applied to a variety of response selection models \cite{ RS1, UMS, BERT-VFT, BERT-FP}. Despite their huge success, existing works still have much room for improvement. 



External knowledge like syntactic information can help the model capture the key fact and get a better grasp of the sentences, thus it is widely used in NLP tasks\cite{ lisa, sla, spider}. 
 LISA \cite{lisa} restricted each token to attend to its syntactic parent in one attention head, SLA \cite{sla} restrained the attention scope based on the distances in the syntactic structure. But these methods focus on intra-sentence syntax, which only consider the syntactic structure in single sentence, ignoring the syntactic interaction across multiple sentences. Moreover, introducing syntax knowledge is less explored in the retrieval-based dialogue task. Recently, SPIDER \cite{spider} proposed the Sentence Backbone Regularization(SBR) task to guide the model to learn the internal relations of subject-verb-object triplets.
Nevertheless, there are several disadvantages in this paradigm: (1) they don't explicitly model the deep syntax information in each sentence (intra-sentence); (2) they also don't take into consideration syntactic dependencies among multiple sentences (inter-sentence), leading to the sub-optimized model performances.

\begin{figure*}[t]
    \vspace{-15pt}
    \centering
    \includegraphics[width=0.80\textwidth]{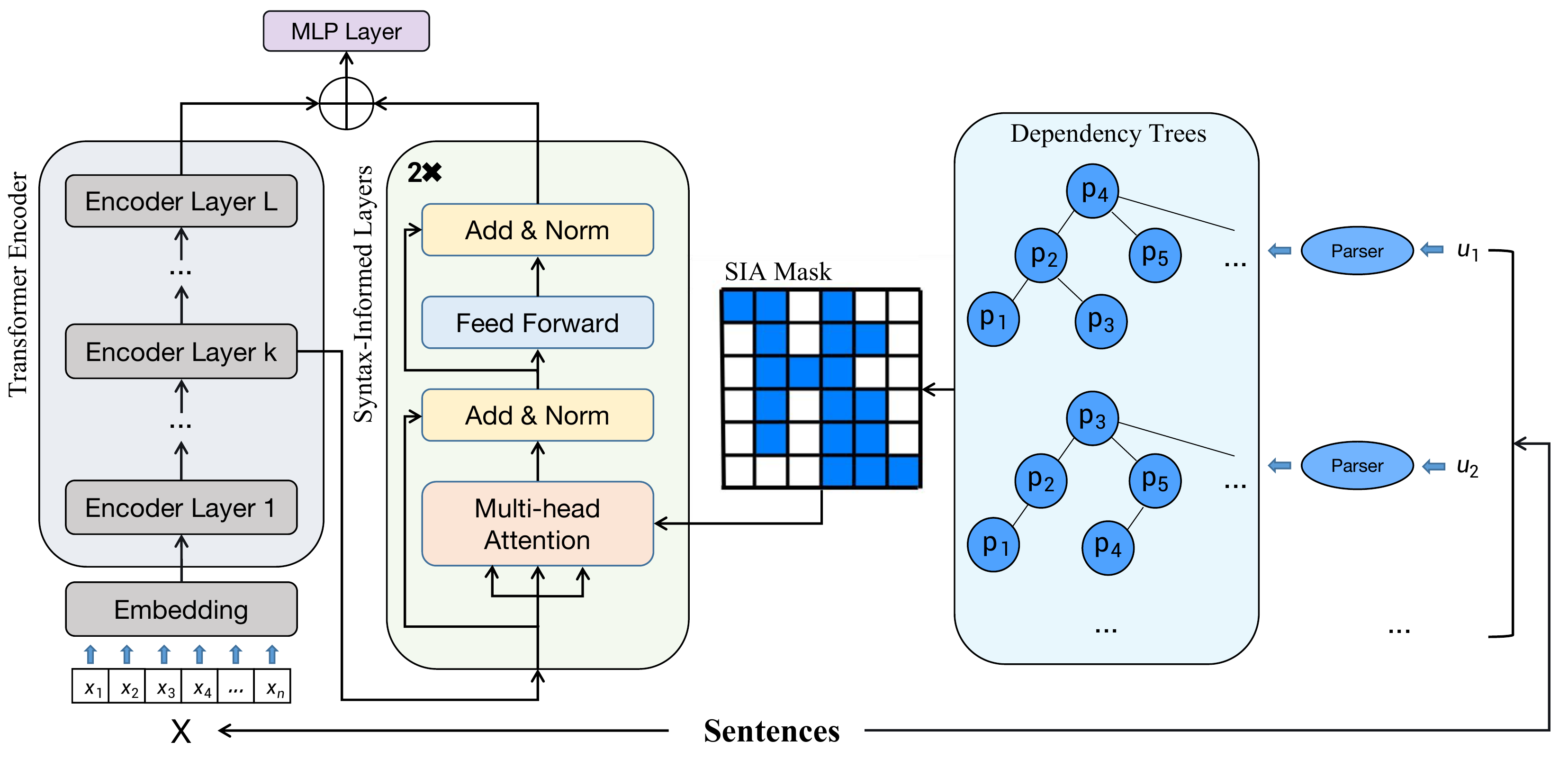}
    \caption{An overview of Syntax-Informed Attention.}
    \label{Fig.1}
\end{figure*}

Inspired by the above concerns, we propose Syntax-Informed Attention (SIA) to fuse the comprehensive syntactic knowledge into current pre-trained language models (PLMs), enhancing the model's understanding of dialogue context. Concretely, SIA consists of two components: the "local" component focuses on modeling local relations of the tokens in intra-sentence dependency tree; the "global" component explicitly considers the inter-sentence syntactic dependencies. To be more specific, SIA is composed of two stacked multi-head self-attention layers, which can be seen as a plug-and-play augmentation on existing Transformer-based models, as shown in Fig.\ref{Fig.1}. Experimentally, we extend our SIA to BERT, the most general model for response selection task. Experiments show that the resulting model achieves consistent performances on three benchmarks, proving the effectiveness and generalizability of SIA.
For example, SIA helps BERT achieves absolute improvements of 1.3\% and 4.5\% in $ R_{10}@1$ on ubuntu and E-commerce, respectively. Further ablations and visualization are conducted to show the interpretability of our method.

Our contributions are summarized as follows: (1) We propose a new attention mechanism to introduce intra- and inter-sentence syntax knowledge into PLMs, named Syntax-Informed Attention (SIA); (2) The proposed SIA can be easily combined with existing Transformer-based models, allowing initialization from pre-trained PLMs with only a few new parameters added; (3) Extensive experiments demonstrate the general superiority of SIA  on three benchmarks.


\section{METHOD}
\label{sec:majhead}


\subsection{Problem Formulation}
\label{ssec:subhead}

Let us consider some dataset $ D={(c_i, r_i, y_i)}_{i=1}^N $ consisting of $ N $ triples, where each triple contains the context $ c_i $, the response $r_i$ and the ground truth $ y_i $. The context $ c_i=\{u_1, u_2, ..., u_M\} $ is a sequence of successive utterances, where $ M $ is the context length. The response $ r_i $ is a single utterance which is potential to be consistent with the context $ c_i $. And $y_i$ is an indicator that takes the value of 1 if that $r_i$ is the proper response for $c_i$, 0 otherwise. The goal of response selection task is to learn a matching model $ g(\cdot, \cdot)$ from the dataset $ D $ that can measure the matching degree $ g(c_i, r_i)$ for any input $ (c_i, r_i) $ pair.

\subsection{Backbone}
\label{ssec:subhead}
Recent works that apply pre-trained language models for response selection frame the task as a context-response sequence binary classification problem. Following previous studies, we select BERT-based models as our baselines. Specifically, the input format of BERT for the response selection is as follows:
\begin{equation}\label{(1)}
    X=[CLS] u_1 [EOU] ...u_M [EOU] [SEP] r [SEP]    
\end{equation}
where $[CLS]$ and $[SEP]$ are the classification symbol and the segment separation symbol of BERT. And $[EOU]$ is a special token located after each utterance in the context that indicates the End of Utterance (EOU). Then we pass the input into the BERT and get the final representations. Following the common settings \cite{BERT-VFT, BERT-FP}, we regard the representation of $[CLS]$ token as the sentence-level representation: $ T_{[CLS]} $.
Then, the $ T_{[CLS]} $ is fed into a simple MLP layer to compute the final matching score $ g(c, r) $ of the input context-response pair:
\begin{equation}\label{(2)}
    g(c, r)=\sigma(W_{[Task]}T_{[CLS]}+b)
\end{equation}
where $\sigma(\cdot)$ stands a sigmoid function, $ W_{[Task]} $ is the task-specific learnable parameter in fine-tuning. Generally, the parameters of BERT will be optimized by the cross-entropy loss.

\subsection{Syntax-Informed Attention}
\label{ssec:subhead}


Although backbone models achieved promising results, they neglect syntax information in dialogue context. In this work, we propose Syntax-Informed Attention, which we call SIA, that can learn syntax-enhanced representations for response selection by making full use of syntactic dependencies. Different from previous works \cite{kasai, syntax_2} that introduce additional syntactic features, we implement SIA by applying attention mask on the self-attention to conduct syntactic restrain. As aforementioned, SIA models both intra- and inter-sentence syntactic information. In the following, we will give detailed information about our methods sequentially.
\begin{spacing}{1.2}
\noindent{\textbf{Dependency Syntactic Parsing}}
\end{spacing}

\begin{figure}[t]
    \includegraphics[width=0.45\textwidth]{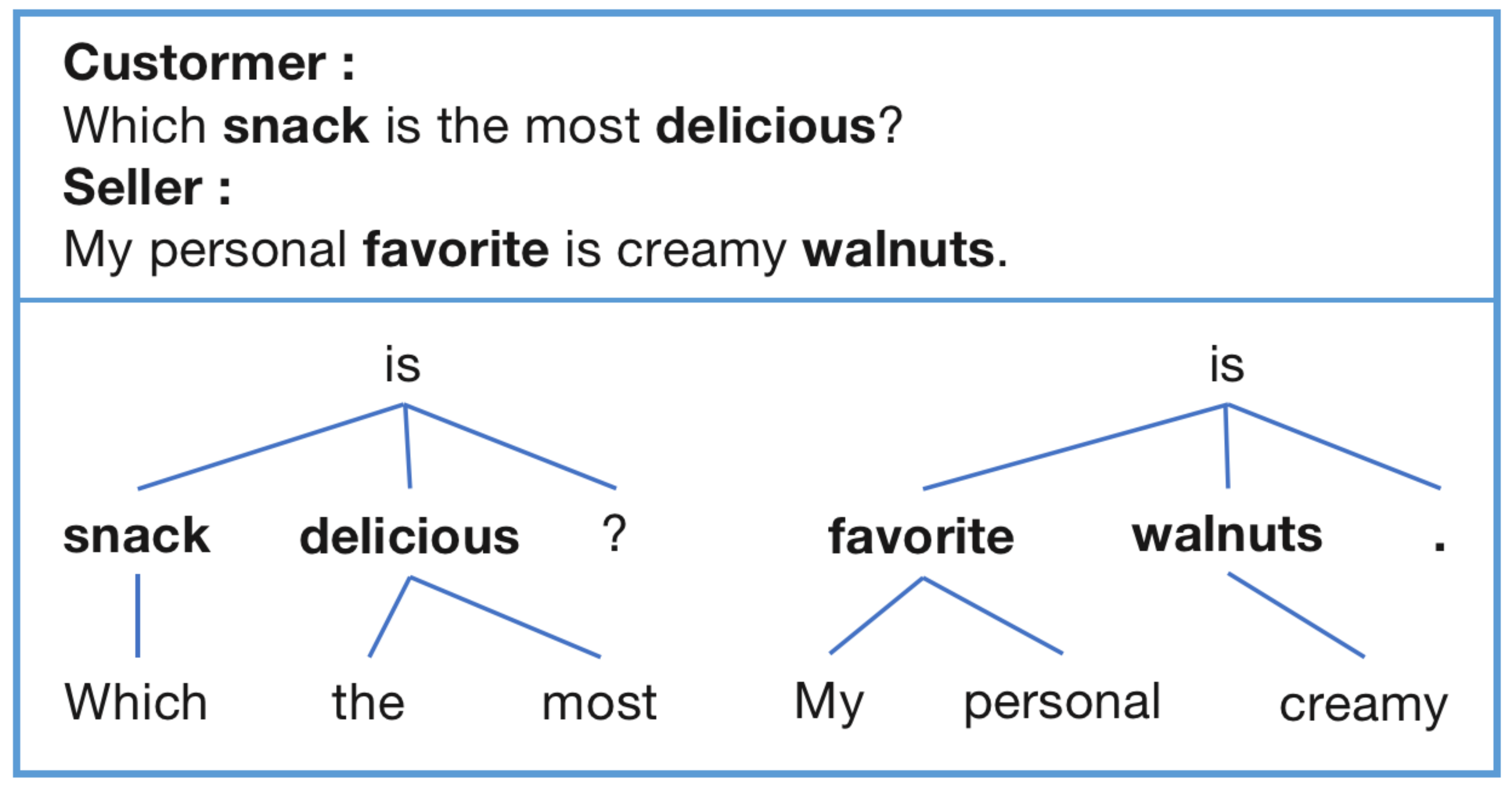}
    \caption{Example of single turn dialogue and dependency trees.} 
    \label{Fig.2}
\end{figure}

In preparation, we need to obtain the syntactic structure. For any single utterance $u_j=\{t_1, t_2, ..., t_L\} $ in the dialogue, we parse and generate its syntactic dependency tree, where the representation of token $  t_i $ will be denoted as $ p_i$. As illustrated in Fig.\ref{Fig.2}, nodes in a dependency tree are connected by asymmetric relationship called dependencies. Specifically, each edge represents a dependency, and the child node depends on the parent node.

\begin{spacing}{1.2}
\noindent{\textbf{Intra-sentence Attention MASK}}
\end{spacing}

The intra-sentence attention mask adapts the idea explored by several works \cite{lisa, sg-net} that a token has strong relation to its syntactic dependency ancestors. Specifically, for any node $p_i$, we denote the set of its syntactic ancestors as $A_i$. To limit attention to only between the node and its syntactic ancestors, the intra-sentence attention mask matrix $ M^{intra} $ can be formulated as follows:

\begin{equation}\label{(5)}
 M^{intra}=\left\{
\begin{array}{rcl}
1, & & {q_j \in A_i \quad or \quad i=j}\\
0, & & {otherwise}\\
\end{array} \right. 
\end{equation}

As seen from the formulation, $ p_j $ can participate in the attention computation of $ p_i $, i.e., $ M^{intra}[i, j]=1 $, only if $ p_j $ is the ancestor of $ p_i $ or $ p_i $ itself. In this way, intra-sentence attention can explicitly model the deep syntax information in each sentence.
\begin{spacing}{1.3}
\noindent{\textbf{Inter-sentence Attention MASK}}
\end{spacing}

The intra-sentence syntax has been widely used in previous work \cite{sg-net, sla}, but this approach of limiting the interaction between tokens to within sentences is far from adequate for response selection since multi-turn dialogue usually contains multiple sentences. This motivates us to find a reasonable mechanism to apply syntactic information extensions to inter-sentence interactions. To this end, we proposed inter-sentence attention mask.

An intuitive idea is to establish interactions between important tokens of each sentence. As illustrated in Fig.\ref{Fig.2}, each child node in the dependency tree is dependent on its parent node, therefore the backbone tokens of a sentence will appear at higher positions of the tree while modifiers will appear at lower positions. According to this property of the dependency tree, an assumption can be made that the depth of $p_i$ in corresponding syntactic tree can roughly represent the importance level of token $ x_i $ in the utterance. 
In order to allow more important words to participate more widely in the attention calculation process of other tokens, the inter-sentence attention mask $ M^{inter} $ is defined as follows:

\begin{equation}\label{(4)}
 M^{inter}=\left\{
\begin{array}{rcl}
1, & & {d_i+d_j \leq m}\\
0, & & {otherwise}\\
\end{array} \right. 
\end{equation}

where $d_i$ is the depth of $p_i$ in the dependency tree. And $m$ is a hyperparameter that defines the maximum range within which upper-level nodes can participate in each other's attention computation.
\begin{spacing}{1.2}
\noindent{\textbf{Syntax-Informed Attention}}
\end{spacing}

To take full advantage of syntactic information, we combine these two constraints in two stacked self-attention layers. Specifically, we implement OR operation on $ M^{inter} $ and $ M^{intra} $ to get $ M^{SIA} $, which will act as the final syntax-informed attention mask. The output from the $ k^{th} $ layer of BERT is projected into distinct key $ K $, value $ V $, and query $ Q $ representations respectively. 
The probing task in \cite{syntax} proves that the intermediate layer of BERT contains relatively richer syntactic information, hence k is set to 6. The syntax-informed attention output calculated as follows:

\begin{equation}\label{(6)}
 W^{SIA}=softmax(\frac{M^{SIA}\cdot(QK^{\top})}{\sqrt{d_k}})V
\end{equation}
where $ d_k $ is the dimension of $ K $. With the syntax-informed attention output $ W^{SIA} $, we can obtain the syntax-informed representations $ H^{SIA} $.

The final output of the model is a combination of syntax-informed representation and the last hidden state of BERT $ H $:
\begin{equation}\label{(7)}
 H'= H+H^{SIA}
\end{equation}
Finally, we fed  $ T_{[CLS]} $ from $ H' $ into a simple MLP and optimize the models with cross-entropy loss.

\begin{table*}[h]
\setlength\tabcolsep{5pt}
\centering
\footnotesize
\caption{Performance comparison on Ubuntu, Douban and E-Commerce datasets}
\begin{tabular}{lcccccccccccccc}
\toprule
\multicolumn{1}{c}{\multirow{2}{*}{\textbf{Models}}} & \multicolumn{3}{c}{\textbf{Ubuntu Corpus}}    & \multicolumn{1}{l}{} & \multicolumn{6}{c}{\textbf{Douban Conversation Corpus}}                                       & \multicolumn{1}{l}{} & \multicolumn{3}{c}{\textbf{E-commerce Corpus}} \\ \cline{2-4} \cline{6-11} \cline{13-15} 
\multicolumn{1}{c}{}                                 & $R_{10}@1$   & $R_{10}@2$   & $R_{10}@5$   & \multicolumn{1}{l}{} & MAP           & MRR           & $P@1$           & $R_{10}@1$   & $R_{10}@2$   & $R_{10}@5$   & \multicolumn{1}{l}{} & $R_{10}@1$    & $R_{10}@2$   & $R_{10}@5$   \\ \toprule
SMN\cite{SMN}                                                  & 72.6          & 84.7          & 96.1          &                      & 52.9         & 56.9          & 39.7          & 23.3          & 39.6          & 72.4          &                      & 45.3           & 65.4          & 88.6          \\
DUA\cite{e-commerce}                                                  & 75.2          & 86.8          & 96.2          &                      & 55.1          & 59.9          & 42.1          & 24.3          & 42.1          & 78.0          &                      & 50.1           & 70.0          & 92.1          \\
DAM\cite{DAM}                                                  & 76.7          & 87.4          & 96.9          &                      & 55.0          & 60.1          & 42.7          & 25.4          & 41.0          & 75.7          &                      & 52.6           & 72.7          & 93.3          \\
IOI\cite{IOI}                                                  & 79.6          & 89.4          & 97.4          &                      & 57.3          & 62.1          & 44.4          & 26.9          & 45.1          & 78.6          &                      & 56.3           & 76.8          & 95.0          \\
RoBERTa-SS-DA\cite{RS1}                                        & 82.6          & 90.9          & 97.8          &                      & 60.2          & 64.6          & 46.0          & 28.0          & 49.5          & 84.7          &                      & 62.7           & 83.5          & 98.0          \\
BERT+SPIDER\cite{spider}                                          & 83.1          & 91.3          & 98.0          &                      & 59.8          & 63.8          & 45.9          & 28.5          & 48.7          & 82.6          &                      & 62.6           & 82.7          & 97.1          \\
$\rm UMS_{BERT}$\cite{UMS}                                                  & 84.3          & 92.0          & 98.2          &                      & 59.7          & 63.9          & 46.6          & 28.5          & 47.1          & 82.9          &                      & 67.4           & 86.1          & 98.0          \\ \toprule
BERT\cite{RS1}                                                 & 80.8          & 89.7          & 97.5          & \multicolumn{1}{l}{} & 59.1          & 66.3          & 45.4          & 28.0          & 47.0          & 82.1          & \multicolumn{1}{l}{} & 61.0           & 81.4          & 97.3          \\
\multicolumn{1}{l}{\textbf{\quad + SIA}}                    & 82.1 & 90.6 & 97.8 & \multicolumn{1}{l}{} & 59.7 & 67.4 & 46.5 & 28.6 & 47.3 & 83.8 & \multicolumn{1}{l}{} & 65.5  & 85.0 & 98.2 \\ \toprule
BERT-FP\cite{BERT-FP}                                                 & 91.1          & 96.2          & \textbf{99.4}          & \multicolumn{1}{l}{} & 64.4          & 68.0          & 51.2          & 32.4          & \textbf{54.2}          & 87.0          & \multicolumn{1}{l}{} & 87.0           & \textbf{95.6}          & 99.3          \\
\multicolumn{1}{l}{\textbf{\quad + SIA}}                    & \textbf{91.4} & \textbf{96.4} & \textbf{99.4} & \multicolumn{1}{l}{} & \textbf{64.8} & \textbf{68.9} & \textbf{52.6} & \textbf{33.3} & \textbf{54.2} & \textbf{87.1} & \multicolumn{1}{l}{} & \textbf{87.4}  & 95.5 & \textbf{99.4} \\ \toprule

\end{tabular}
\label{table1}
\end{table*}
\section{EXPERIMENTS}
\label{sec:majhead}

\subsection{Dataset and Experimental Settings}
\label{ssec:subhead}

\begin{spacing}{1.2}
\noindent{\textbf{Dataset}}
\end{spacing}

 We evaluate our method on three widely used response selection benchmarks: Ubuntu Corpus V1\cite{ubuntu} is an English dialogue corpus, which is mainly about how to troubleshoot the Ubuntu OS, Douban Conversation Corpus\cite{douban} is a Chinese open-domain dialogue corpus, and  E-Commerce Corpus\cite{e-commerce} is a Chinese dialogue corpus collected from Taobao, the largest e-commerce platform in China.




\begin{spacing}{1.3}
\noindent{\textbf{Experimental settings}}
\end{spacing}

To demonstrate the effectiveness and generality of our proposed SIA, we extend it on BERT and BERT-FP, the latter is the state-of-the-art for multi-turn response selection. And we evaluated our model by several retrieval metrics. Firstly, we figured out the proportion of answers in top-k ranked candidates, denoted as $ R_n@k(k={1, 2, 5}) $. And then, three other metrics [mean average precision (MAP), mean reciprocal rank (MRR), and precision at one (P@1)] are used especially for Douban, as in this case there can be more than one positive response in candidates. Our dependency parser is taken from spaCy.

\begin{spacing}{1.3}
\noindent{\textbf{Baseline models}}
\end{spacing}
We compared our method with some previous works. In a nutshell, SMN\cite{SMN}, DUA\cite{e-commerce}, DAM\cite{DAM} and IOI\cite{IOI} are non-PLM-based models. RoBERTa-SS-DA\cite{RS1} discriminates speakers by speaker segmentation and applies dialogue augmentation. SPIDER\cite{spider} proposes two additional training objectives to capture dialogue-exclusive features. UMS\cite{UMS} is a multi-task framework consisting of three auxiliary tasks. BERT-FP applies a fine-grained post-training and is the current state-of-the-art of response selection.

\subsection{Results and Discussion}
\label{ssec:subhead}
\begin{spacing}{1.3}
\noindent{\textbf{Quantitative Results}}
\end{spacing}

Table \ref{table1} shows the quantitative results on Ubuntu, Douban, and E-commerce datasets, all the experimental results of baselines are cited from previous works. Referring to the table, SIA significantly improves the two baseline in all of the metrics on the three datasets, which helps BERT performs better than RoBERTa-SS-DA\cite{RS1} and BERT+SPIDER\cite{spider} on most evaluation criteria. Specifically, SIA improved the performance of BERT by 1.3\% and 4.5\% in $ R_{10}@1$ on Ubuntu and E-Commerce. The extensive experiment results prove the effectiveness of our proposed SIA.
\begin{table}[h]
\vspace{-10pt}
\footnotesize
\caption{Ablation study on E-commerce}
\centering
\begin{tabular}{rccc}
\toprule
\multicolumn{1}{c}{\multirow{2}{*}{\textbf{Models}}} & \multicolumn{3}{c}{\textbf{E-commerce Corpus}} \\ \cline{2-4} 
                                 & $R_{10}@1$          & $R_{10}@2$         & $R_{10}@5$         \\ \toprule
\multicolumn{1}{l}{BERT}                             & 61.0           & 81.4          & 97.3          \\
\multicolumn{1}{l}{\quad +vanilla attention}               & 61.6           & 79.9          & 95.8          \\
\multicolumn{1}{l}{\quad+intra-}                          & 62.3           & 82.5          & 97.4          \\
\multicolumn{1}{l}{\quad+inter-}                          & 65.1           & 83.3          & 97.9          \\
\multicolumn{1}{l}{\textbf{\quad+SIA}}                    & \textbf{65.5}  & \textbf{85.0} & \textbf{98.2} \\ \toprule
\label{tabel2}
\end{tabular}
\vspace{-10pt}
\end{table}

\begin{spacing}{1.3}
\noindent{\textbf{Ablation Study}}
\end{spacing}

We performed ablation studies on the E-commerce Corpus to assess the effectiveness of our method. As shown in Table \ref{tabel2}, we applied four settings separately on the BERT. It can be found out that vanilla attention performs the worst and even causes performance deterioration on two evaluation metrics. While both inter- and intra- improve the model performance, inter- shows an obvious advantage, further speaks for the usefulness of inter-sentence attention mask. The best result is achieved when combining both intra- and inter-sentence attention mask with the response selection criterion using our representation fusion method.

\begin{figure}[]
    \centering
    \includegraphics[width=0.48\textwidth]{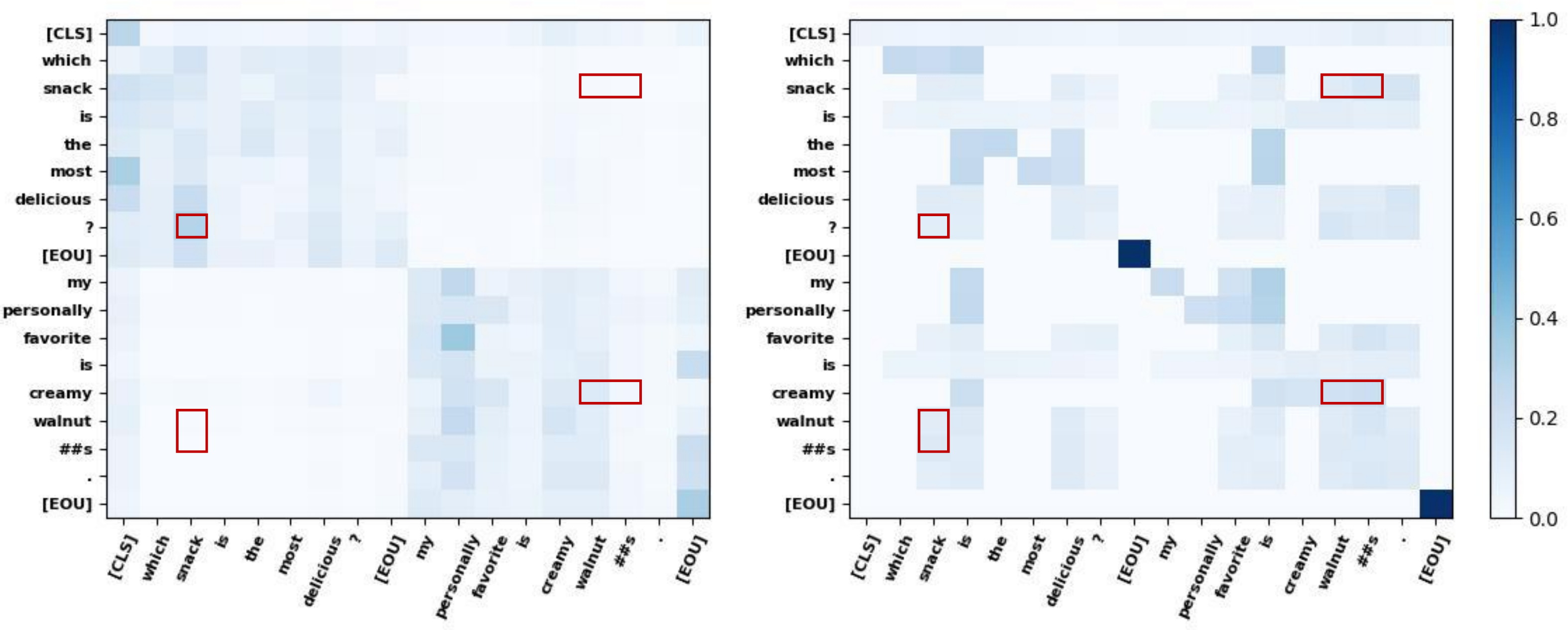}
    \caption{Visualization of vanilla attention score (left) and Syntax-Informed attention score (right).}
    \label{Fig.3}
\end{figure}

\begin{spacing}{1.2}
\noindent{\textbf{Visualization}}
\end{spacing}

As shown in Fig.\ref{Fig.3}, we illustrate how our SIA works by comparing the distributions of the vanilla attention score of BERT and our proposed SIA. We can find high attention scores between less related tokens such as "?" and "delicious" in the vanilla attention, but not in SIA. And under the constraint of inter-sentence attention, "walnuts" heavily attend to "snack", while under the constraint of intra-sentence attention, "walnuts" heavily attend to "creamy".

\section{Conclusion}
\label{sec:majhead}

In this paper, we propose Syntax-Informed Attention for multi-turn response selection, which introduces intra- and inter-sentence syntax to the models for helping them better understand the informative multi-turn dialogue. Experimental analyses on three public benchmarks verify the generality and effectiveness of our method. In the future, we will take into account the category of dependencies, i.e., use more fine-grained syntactic information to assist the model.


\bibliographystyle{IEEEbib}
\bibliography{strings,refs}

\end{document}